# Multimodal large language model for wheat breeding: a new exploration of smart breeding

Guofeng Yang[a], Yu Li[b], Yong He[a], Zhenjiang Zhou[a], Lingzhen Ye[c], Hui Fang[a], Yiqi Luo[d], Xuping Feng[a]*

**Affiliations**

[a] *College of Biosystems Engineering and Food Science, Zhejiang University, Hangzhou 310058, Zhejiang, China*

[b] *Zhejiang Society of Agricultural Machinery, Hangzhou 310003, Zhejiang, China*

[c] *College of Agriculture and Biotechnology, Zhejiang University, Hangzhou 310058, Zhejiang, China*

[d] *Soil and Crop Sciences Section, School of Integrative Plant Science, Cornell University, Ithaca 14853, New York, USA*

*Correspondence: fengxp@zju.edu.cn

**Abstract**

UAV remote sensing technology has become a key technology in crop breeding, which can achieve high-throughput and non-destructive collection of crop phenotyping data. However, the multidisciplinary nature of breeding has brought technical barriers and efficiency challenges to knowledge mining. Therefore, it is important to develop a smart breeding goal tool to mine cross-domain multimodal data. Based on different pre-trained open-source multimodal large language models (MLLMs) (e.g., Qwen-VL, InternVL, Deepseek-VL), this study used supervised fine-tuning (SFT), retrieval-augmented generation (RAG), and reinforcement learning from human feedback (RLHF) technologies to inject cross-domain knowledge into MLLMs, thereby constructing multiple multimodal large language models for wheat breeding (WBLMs). The above WBLMs were evaluated using the newly created evaluation benchmark in this study. The results showed that the WBLM constructed using SFT, RAG and RLHF technologies and InternVL2-8B has leading performance. Then, subsequent experiments were conducted using the WBLM. Ablation experiments indicated that the combination of SFT, RAG, and RLHF technologies can improve the overall generation performance, enhance the generated quality, balance the timeliness and adaptability of the generated answer, and reduce hallucinations and biases. The WBLM performed best in wheat yield prediction using cross-domain data (remote sensing, phenotyping, weather, germplasm) simultaneously, with $R^2$ and RMSE of 0.821 and 489.254 kg/ha, respectively. Furthermore, the WBLM can generate professional decision support answers for phenotyping estimation, environmental stress assessment, target germplasm screening, cultivation technique recommendation, and seed price query tasks. This study aims to provide intelligent and integrated solutions for wheat breeding goals, to help breeding work be carried out efficiently, to accelerate the breeding process of excellent varieties, and to provide scientific basis and technical support for achieving sustainable agricultural development and ensuring food security.

## 1. Introduction

As one of the basic food crops for mankind, wheat breeding is facing unprecedented challenges and opportunities due to the global food crisis and the promotion of sustainable agricultural development (Cheng et al., 2024; Senapati et al., 2022). Although traditional breeding methods have achieved remarkable results, their efficiency and accuracy have gradually revealed their limitations in the face of complex and changing climatic conditions, increasingly serious threats from pests and diseases, and constantly upgrading consumer demands (Hu et al., 2022; Xiong et al., 2021). Due to the diversity of wheat varieties, breeding information has long lacked a unified tool, and data knowledge has shown an "isolated" distribution, which has created barriers to the learning of wheat breeding knowledge (Bhat et al., 2023). Simultaneously, since wheat breeding involves the intersection of multiple disciplines such as biology, genetics, weather, and soil science, professionals have to cross literature and data from many fields when engaged in breeding work, and even need to write code to access data, which greatly limits their work efficiency (Kaur et al., 2021). Smart breeding, as an innovative mode that integrates modern information technology, biotechnology, and agricultural science, is gradually becoming a key path to solving this problem (H. Li et al., 2024; Xu et al., 2022).

With the rapid development of unmanned aerial vehicle (UAV) technology, its application in the agricultural field has become more and more extensive, especially UAVs have shown great potential in intelligent breeding (Das et al., 2021; Fei et al., 2023; Jiang et al., 2021). With its advantages of high efficiency, precision and flexibility, it is becoming an indispensable tool in smart breeding and can provide important monitoring data for smart breeding. Image recognition technology can accurately extract wheat growth indicators from RS images, such as chlorophyll content (Feng et al., 2023; Zhang et al., 2024) and leaf area index (Chen et al., 2022; Du et al., 2023); text analysis technology can extract valuable germplasm descriptions (L. Liu et al., 2024; Sansaloni et al., 2020), breeding optimization strategies (J. Xu et al., 2023; Yao et al., 2022) and market trends (Garg et al., 2022; Padhy et al., 2024) from research manuscripts, reports and databases. The complementarity of image and text information not only enhances the credibility and interpretability of the data, but also promotes the integration of interdisciplinary knowledge, opening up new avenues for wheat breeding research. At present, there is an urgent need to integrate cross-domain data (RS data, phenotyping data, environment data, germplasm data, cultivation data, and price data), use artificial intelligence algorithms and big data technology to build a cross-scale and cross-domain comprehensive analysis tool for wheat breeding, accelerate the screening and optimization of wheat target varieties, and improve the efficiency and accuracy of wheat breeding.

In recent years, the large language model (LLM) has attracted much attention due to their powerful text generation and understanding capabilities (X. Huang et al., 2024; Lin et al., 2024). After fine-tuning, the LLM has shown strong interactive capabilities and the potential to improve productivity (Min et al., 2024). However, since the LLM can only process plain text and cannot process images, voice, and video, their application scope is limited (Wang et al., 2023). In the context of cross-domain, multi-modal data fusion applications, several multimodal large language models (MLLM) have been developed to enhance the ability of LLM to perceive and understand visual signals (AI et al., 2024, p. 0; Bai et al., 2023; Z. Chen et al., 2024; DeepSeek-AI et al., 2024; GLM et al., 2024; F. Li et al., 2024). Although a lot of study has been done to explore the limitations and effectiveness of MLLM, the current open-source general MLLM still has the problem of insufficient accuracy in professional applications (D. Huang et al., 2024; C. Li et al., 2024), such as wheat breeding selection. When faced with wheat breeding query, general MLLM often evades the question or gives irrelevant answers due to lack of breeding knowledge. This hinders the further exploration and application of MLLM in the field of wheat breeding.

However, the research and application of MLLM in crop breeding is still in its early stages and faces many challenges and opportunities (Zhu et al., 2024). The MLLM needs to integrate more complex data from different sources, including but not limited to RS data, phenotyping data, and environment data during the crop growth period. Data from these sources vary significantly in format, dimensionality, and sparsity. How to efficiently and accurately integrate these cross-domain heterogeneous data is a difficulty in current research (Kuska et al., 2024). MLLM needs to have strong learning and generalization capabilities to adapt to the needs of smart breeding and predict the performance of crops under different environmental conditions, which places extremely high demands on the model's algorithm design and computing resources. Additionally, professional knowledge and experience in the field of crop breeding are crucial for the development and application of the model (J. Li et al., 2024). How to integrate professional knowledge into model design to make the model more in line with actual breeding needs, while discovering new breeding strategies and optimization solutions through the model, is another issue that needs in-depth exploration. MLLM is expected to reveal the deep mechanisms of crop trait formation by integrating cross-modal information and accelerate the screening and breeding process of new varieties.

This study aims to innovatively construct a MLLM for wheat breeding (WBLM) through cross-domain data fusion and cutting-edge technology application, and explore its potential in wheat breeding goals. The purposes of this study are to: (i) evaluate the contribution of the integrated application of domain knowledge technologies (supervised fine-tuning, retrieval augmented generation, and reinforcement learning from human feedback) to achieving wheat breeding goals, and analyze the performance of cross-domain data fusion in wheat yield prediction; (ii) explore the response of WBLM in coping with multidimensional breeding goals, and generate personalized decision support from the aspects of phenotyping estimation, environmental stress assessment, target germplasm screening, cultivation technique recommendation, seed price query; (iii) release the study dataset to promote research and application innovation in this field.

## 2. Study area and data

*2.1 Experimental sites and design*

The experiment was conducted for two years (2021-2022 and 2022-2023), with 305 and 351 varieties, respectively. The varieties in the second year included the varieties in the first year and 46 varieties were added (Table S1). The experimental sites were set up in Changxing and Yuhang Agricultural Experimental Bases of Zhejiang University Agricultural Experiment Station in Zhejiang Province, China (Fig.1a, Table S2). Specific planting information is shown in Table S3. The experimental fields were not sprayed with pesticides and herbicides, and weeds were not controlled manually or mechanically. Maintaining the natural growth environment is to develop competitive wheat varieties that have a greater competitive advantage than wheat weeds. The water required for wheat growth comes from natural rainfall. Both places have a subtropical monsoon climate.

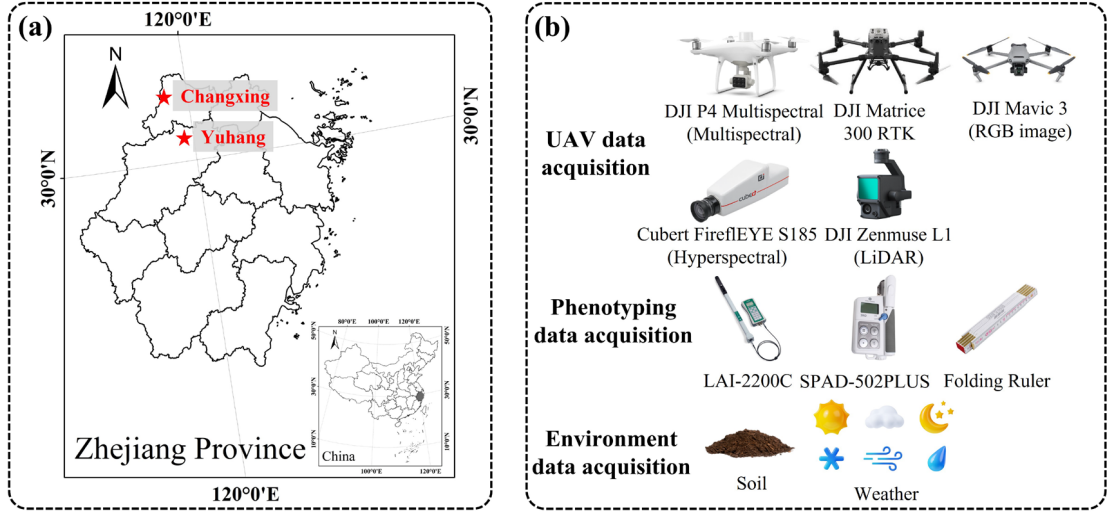

Fig. 1. (a) Experimental sites. (b) Multi-source data acquisition.

*2.2. Data acquisition*

*2.2.1. UAV data*

For the experimental field, we used multiple sensors and multiple UAVs to acquire hyperspectral (HS), lidar, multispectral (MS), and RGB (red, green, blue) data on the same day and compiled them into a RS dataset (Fig. 1b). The Matrice 300 RTK (DJI, China) was selected as the flight platform to carry the Zenmuse L1 (DJI, China) lidar camera and the FireflEYE S185 (Cubert, Germany) hyperspectral camera to obtain lidar data and HS data respectively. FireflEYE S185 covers the visible and near-infrared bands from 450 to 950 nm, has 125 spectral channels, and can achieve synchronous frame imaging, spectral resolution: 8 nm @ 532 nm, spectral sampling interval 4 nm. Zenmuse L1 has a range of 190m @ 10%, 100 klx, point cloud data rate multi-echo: maximum 480,000 points/second, elevation accuracy: 5cm @ 50m, plane accuracy: 10cm @ 50m. The Phantom 4 multispectral (DJI, China) and the Mavic 3E (DJI, China) were used to obtain MS data and RGB image respectively. The MS includes five bands, blue: 450 nm ± 16 nm; green: 560 nm ± 16 nm; red: 650 nm ± 16 nm; red edge: 730 nm ± 16 nm; near infrared: 840 nm ± 26 nm. The Mavic 3E is equipped with a wide-angle camera 4/3 CMOS, with 20 million effective pixels and a maximum photo size of 5280×3956. The telephoto camera has an equivalent focal length of 162 mm, 12 million pixels, and 56x hybrid zoom.

UAVs were used to collect data during the wheat growth period many times, all under low wind speed and clear and cloudless conditions at noon using different UAVs and sensors. During automatic data collection, the UAV's flight altitude was 15 m, and other flight parameters such as overlap rate were set according to the parameters recommended by the equipment manufacturer. During manual data collection, the Mavic 3E UAV flight altitude was maintained at 3 m and a 7x zoom was set to obtain images and video data of the experimental field. In particular, since the UAV used supports image-free control technology and the experimental field area is relatively small, no ground control points were set and no geometric correction and registration of the data was performed.

*2.2.2. Phenotypic data*

Phenotypic data collection involves chlorophyll content, leaf area index (LAI), canopy height (CH) and yield (Fig. 1b). Field phenotyping data of wheat were collected six times during 2021-2022 (on January 19, March 8, April 2, April 15, April 29, and May 18, 2022) and nine times during 2022-2023 (on December 27, 2022; February 16, March 4, March 19, April 7, April 22, May 1, May 13, May 19, 2023). UAV RS and phenotyping data were collected on the same day.

The chlorophyll content of the wheat canopy was measured using a SPAD-502 PLUS chlorophyll meter (Konica Minolta, Japan). Five wheat plants with representative growth in each wheat planting plot of the field experiment were randomly selected. For each wheat plant, three leaves without pests and diseases, physiological spots, and mechanical damage were selected, and the tip, middle, and base of each leaf were measured. All the measured values were averaged as the SPAD value of the canopy leaves of the wheat germplasm in the plot. The LAI of the wheat canopy in each plot was measured using a LAI-2000C plant canopy analyzer (LI-COR, USA). Five sampling points with uniform growth were selected at the four corners and the center of each plot, and the LAI values were measured and recorded using a LAI-2200C according to the standard method. Each sampling point was measured three times and the average value was taken as the LAI of the sampling point. We recorded the average of the LAI of the five sampling points after scatter correction as the LAI of the plot. In each plot, five CH samples were randomly selected for measurement using a tape measure, and the average value was taken as the CH of the plot. The manually measured CH was used to verify the CH derived from the UAV light detection and ranging (lidar) point cloud. Wheat was harvested when mature in each plot. After threshing, the grain from each yield plot was weighed and yield was expressed as kg ha$^{-1}$ and standardized to a moisture content of 12.5% (State Administration for Market Regulation and Standardization Administration of the P.R.C, 2023).

*2.2.3. Environmental data*

The weather data continuously recorded by meteorological equipment at each agricultural experimental base and the meteorological observation station of the China Meteorological Administration were used to provide us with weather data for two-year growing season (Fig. 1b). The data mainly include daily average temperature, dew point temperature, precipitation, net solar radiation intensity, wind speed and other data. The average annual temperature is 15.6 °C, the average relative humidity is 76%, the annual rainfall is 1309 mm, and the average of 1810 hours of sunlight per year at Changxing Agricultural Experiment Station. The average annual temperature is 16.2 °C, the average relative humidity is 68%, the annual rainfall is 1400 mm, and the average of 1970 hours of sunlight per year at Yuhang Agricultural Experiment Station.

The test soil was collected before sowing at different locations (0-20 cm) in the experimental field. The collected soil was placed in turnover boxes and then sent for testing in time to determine the physical and chemical properties of the soil. Soil pH = 6.2, total N = 1.32 g kg$^{-1}$, available potassium = 94.9 mg kg$^{-1}$, available phosphorus = 1.9 mg kg$^{-1}$ and soil organic C = 12.4 g kg$^{-1}$ at Changxing Agricultural Experiment Station. Soil pH = 6.14, total N = 1.68 g kg$^{-1}$, available potassium = 187 mg kg$^{-1}$, available phosphorus = 63.6 mg kg$^{-1}$ and soil organic C = 16 g kg$^{-1}$ at Yuhang Agricultural Experiment Station.

*2.3. Data processing*

*2.3.1. Data preprocessing*

The acquired UAV data were preprocessed to generate digital orthophoto map (DOM) for subsequent processing according to each plot. Hyperspectral data were converted into reflectance images after radiometric correction using the supporting Cubert Utils Touch software (Cubert, Germany) and radiometric correction plates. Then, Agisoft Metashape (Agisoft, Russia) was used to align photos, create dense point clouds, generate grids, and generate textures to get DOM for subsequent processing. The MS data was imported into Pix4Dmapper (Pix4D, Switzerland) software for initialization processing, point cloud and texture, DSM and DOM generation operations. The irradiance value captured by the light intensity sensor during flight was used to compensate the MS bands for illumination, eliminating the interference of ambient light on data collection. The bands are then radiometrically calibrated using the radiometric calibration plate image acquired simultaneously during data collection. Furthermore, the radiometrically calibrated images were stitched to generate DOM. Finally, the DOM was normalized to obtain reflectance images for subsequent generation of vegetation indexes (VI). The lidar data were processed using DJI Terra (DJI, China) to generate three-dimensional point clouds. The RGB images were imported into Pix4Dmapper to stitch RGB DOM. In addition, one frame of the video obtained by manual flight is extracted every five seconds as an image for storage. These images were divided according to different germplasms (different plots). Then, the wheat heads (WH) in the images were manually annotated to produce a WH dataset.

*2.3.2. Spectral data processing*

The processed MS and HS bands were used for VIs calculation to analyze the canopy spectral features of different plots. VIs commonly used for phenotyping estimation and grain yield prediction were selected, including NDVI, SAVI, kNDVI, NIRv and PSRI (Table 1).

The fractional vegetation cover (FVC) of the plots, i.e., the percentage of the vegetation area to the plot area (Yang et al., 2022), was calculated as a valuable indicator of crop density and structural information. The vegetation area of the plot was extracted from the MS image by excluding the background soil using a Transformer-based segmentation algorithm, similar to related study (Cui et al., 2023). The number of vegetation pixels in each plot was then divided by the total number of pixels in that plot to calculate FVC (Maimaitijiang et al., 2020).

*2.3.3. Lidar data processing*

The CH was extracted from LiDAR point clouds and used as canopy structure features. The bare ground digital elevation model (DEM) was created using point cloud data acquired before the emergence of seedlings in the field. Then, the digital surface model (DSM) representing all objects (vegetation) on the ground was constructed based on the point cloud data acquired at different times. Thus, the CH is obtained by pixel-level subtraction of the generated DSM and DEM (Maimaitijiang et al., 2020). Finally, the obtained lidar data were used to extract CH in different plots in the wheat field.

In addition, the measurement tool of Terra software (DJI, China) was used to mark the reference surface for canopy volume (CV) measurement of each plot based on the lidar point cloud and the boundaries of different plots. The sum of the excavated volume and the filled volume above and below the reference surface is taken as the spatial volume of the reference surface. It should be noted that since there are two types of reference planes, the lowest point (the plane on which it is located) and the average plane, the average value of the volumes obtained from the lowest point and the average plane is taken as the spatial volume of the plot (Table 1).

*2.3.4. RGB image processing*

To obtain the lodging level of different plots in the wheat field, we refer to a study to construct a wheat lodging area segmentation model (Zhang et al., 2023). Then we use the ratio of the lodging area extracted from the plot to the plot area to determine the plant lodging (PL) level of the plot. In our experiment, the lodging levels were divided into: no lodging (0), slight lodging (0-50%), severe lodging (50-100%), and special (no crop or only a few plants) (Saskatchewan Seed Growers' Association, 2024).

We implemented the detection and counting of WH in the field based on the two-stage method FR-Transformer proposed in our previous study (Zhu et al., 2022) and

RGB images. Then, we used this method to obtain the number of WH in the preprocessed WH images of each plot, and averaged the number of WH in each plot. Finally, the ground coverage area of the WH image was calculated using the UAV flight altitude and camera field of view (Avola et al., 2021), and the number of WH per unit area of the plot was further calculated.

Due to the lack of herbicide spraying and manual weeding, weeds in wheat fields grow randomly and in large numbers. To study the environmental (weed) stress of wheat growth, we focused on the severity of weeds in each plot and the area 10 to 20 cm wide outside the plot in the wheat field. We referred to a study (Anderegg et al., 2023) to extract the number of pixels in the plot and specific area and the number of weed pixels in them, and then used the ratio of the number of weed pixels to the number of pixels in the plot and specific area to classify each plot according to the weed level (WL). The weed level can be roughly divided into: no weeds (0-10%), slight weeds (10-40%), moderate weeds (40-70%), and severe weeds (70-100%).

Table 1. Definition of features extracted from different sensors

| Sensor | Features | Formulation | References |
|---|---|---|---|
| MS (Spectral information) | Normalized Difference Vegetation Index | $NDVI = \frac{(NIR - R)}{(NIR + R)}$ | (Tucker, 1979) |
| | Soil-Adjusted Vegetation Index | $SAVI = \frac{(1 + L) \times (NIR - R)}{(NIR + R + L)}$, L = 0.5 | (Huete, 1988; Richardson and Everitt, 1992) |
| | kernel Normalized Difference Vegetation Index | $kNDVI = tanh\left(\left(\frac{NIR-R}{2\sigma}\right)^2\right)$, where $\sigma$ is a tunable length-scale parameter intended to capture nonlinear sensitivity of NDVI to vegetation density. If $\sigma = 0.5(NIR + R)$, which simplifies to $kNDVI = tanh((NDVI)^2)$. | (Camps-Valls et al., 2021) |
| | Near-Infrared Reflectance of Vegetation | $NIR_v = NIR \times NDVI$ | (Badgley et al., 2017) |
| | Plant Senescence Reflectance Index | $PSRI = \frac{(R-G)}{NIR}$, $R$, $G$, and $NIR$ denote the reflectance of the red, green, and near-infrared bands from the 5-band Phantom 4 multispectral, respectively. | (Cao et al., 2019) |
| HS (Spectral information) | Normalized Difference Vegetation Index | $NDVI = \frac{(NIR-R)}{(NIR+R)}$ | (Tucker, 1979) |
| | Soil-Adjusted Vegetation Index | $SAVI = \frac{(1+L)\times(NIR-R)}{(NIR+R+L)}$, $L = 0.5$ | (Huete, 1988; Richardson and Everitt, 1992) |
| | kernel Normalized Difference Vegetation Index | $kNDVI = tanh\left(\left(\frac{NIR-R}{2\sigma}\right)^2\right)$, where $\sigma$ is a tunable length-scale parameter intended to capture nonlinear sensitivity of NDVI to vegetation density. If $\sigma = 0.5(NIR+R)$, which simplifies to $kNDVI = tanh((NDVI)^2)$. | (Camps-Valls et al., 2021) |
| | Near-Infrared Reflectance of Vegetation | $NIR_v = NIR \times NDVI$ | (Badgley et al., 2017) |
| | Plant Senescence Reflectance Index | $PSRI = \frac{(R_{680}-R_{500})}{R_{750}}$, $R_{680}$, $R_{500}$ and $R_{750}$ denote the reflectance at 680 nm, 500 nm, and 750 nm respectively. | (Cao et al., 2019) |
| LiDAR (Structural information) | Canopy Height (CH) | CH = DSM - DEM | (Maimaitijiang et al., 2020) |
| LiDAR (Structural information) | Canopy Volume (CV) | CV = Excavated volume + Filled volume, DJI Terra software (DJI) | https://enterprise.dji.com/dji-terra |
| RGB image (Structural information) | Plant Lodging (PL) | $PL = \frac{\text{Number of lodging pixels in the plot}}{\text{Total number of the plot pixels}} \times 100\%$ | (Zhang et al., 2023) |
| RGB video (Reproduction information) | Wheat Head (WH) | WH = The number of wheat head detection | (Zhu et al., 2022) |
| RGB image (Environmental information) | Weed Level (WL) | $WL = \frac{\text{Number of weed pixels in the plot and specific area}}{\text{Total number of the plot and specific area pixels}} \times 100\%$ | (Anderegg et al., 2023) |
| MS (Environmental information) | Fractional Vegetation Coverage (FVC) | $FVC = \frac{\text{Number of crop pixels in the plot}}{\text{Total number of the plot pixels}} \times 100\%$ | (Cui et al., 2023) |

\* *R, G* and *NIR* are the pixel values of the red, green and near-infrared bands, respectively.

## *2.4. Construction of Cross-domain knowledge base*

The cross-domain knowledge base (in Chinese and English) consists of multi-source datasets in the field (Fig. 2a) and external domain knowledge base (Fig. 2b). The multi-source datasets in the field include UAV RS data, phenotyping data, weather data. The external domain knowledge base includes:

(1) wheat germplasm data (22k), mainly including variety name, place of origin, nutritional quality, resistance and agricultural traits. In terms of nutritional quality, key parameters such as crude protein, lysine, and sedimentation value are involved. In terms of resistance, it covers important characteristics such as disease resistance (stripe rust, leaf rust, powdery mildew, etc.), drought resistance, and cold resistance. In addition, agronomic traits such as maturity, plant height, thousand grain weight, and grain hardness are also recorded.

(2) wheat cultivation technique data (10k), covers the entire process from pre-planting preparation to post-harvest storage, mainly including the selection of suitable varieties, soil preparation, sowing, fertilizer management, irrigation and pest and disease control techniques.

(3) wheat plant protection technique data (10k), mainly including disease prevention and control, pest prevention and control, weed management, as well as new techniques such as precision plant protection based on artificial intelligence and RS, agricultural UAV flight control and green prevention and control.

(4) wheat seed price data (20k), mainly including observation point, variety name, price, specification, planting area and time, etc. For observation points, the economic level and planting demand of different regions lead to price differences; for varieties, different varieties have different characteristics and application scenarios, and high-quality varieties are usually more expensive; different specifications of packaging and planting areas affect the final selling price; prices change over time due to changes in market supply and demand, especially natural disasters or policy adjustments.

The wheat germplasm data comes from the Chinese Crop Germplasm Information Network (https://www.cgris.net/). The data on wheat cultivation technique and wheat plant protection technique come from search engines (Google, Bing, Baidu), and the search terms include "wheat cultivation technique, wheat plant protection technique, 小麦栽培技术, 小麦植保技术". The wheat seed price data comes from the National Seed Market Monitoring Information Release Platform (http://202.127.45.18/). All data from this study have been made publicly available (https://doi.org/10.5281/zenodo.13736773).

## 3. Methods

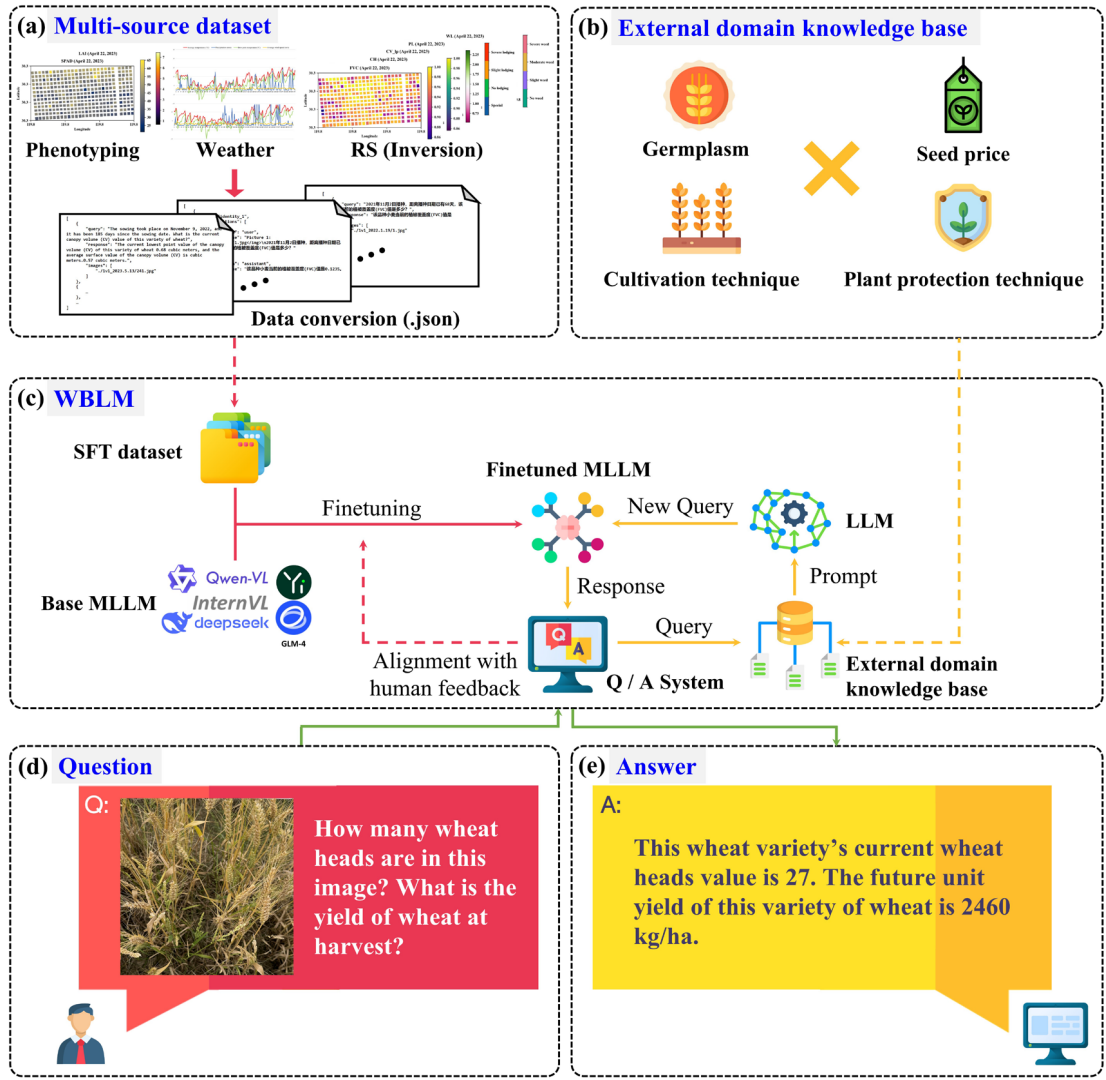

Fig. 2. A workflow for the construction and application of a multimodal large language model for wheat breeding (WBLM). (a) Multi-source dataset construction. (b) External domain knowledge base construction. (c) The WBLM with domain knowledge is constructed using supervised fine-tuning, retrieval augmented generation, and reinforcement learning from human feedback. (d) The user's question (image and text) is sent to WBLM. (e) The WBLM answers the question.

*3.1. Construction of WBLM*

We construct a WBLM with domain knowledge using the supervised fine-tuning (SFT), retrieval augmented generation (RAG) and reinforcement learning from human feedback (RLHF) technologies based on different MLLMs and cross-domain knowledge base. WBLM consists of two parts. The part 1 builds a RAG system based on the LlamaIndex large model application framework (https://www.llamaindex.ai/) and the Milvus vector database (https://milvus.io/) (Fig. 2c, yellow line). The part 2 uses STF and RLHF methods to obtain a WBLM that follows breeding preferences (Fig. 2c, red line). We regard the input of part 1 as a query (Fig. 2d), and the original query and the output of part 1 as a new query. The part 2 receives the new query, processes it, and then outputs the result to the Q/A system (Fig. 2e).

To make WBLM better adapt to breeding tasks, we use RAG technology to conveniently and efficiently supplement the knowledge not covered by STF and RLHF methods. First, we convert the data of the files in the external domain knowledge base, use BGE-M3 to generate embedding vectors (J. Chen et al., 2024), and then import the data into the Milvus vector database. Then, LlamaIndex receives user input and initiates a retrieval request to the Milvus database based on the input, using a hybrid retrieval (BGE-M3) and re-ranking (BGE-Reranker-V2-M3) pipeline. LlamaIndex merges the retrieval results with the input to form a new prompt. After that, LlamaIndex inputs the new prompt to InternLM2.5-7B-Chat to achieve reasoning and answer generation based on the retrieved knowledge (Cai et al., 2024).

The training process using the STF and RLHF methods is divided into three stages. In the first stage, we collect demonstration data and train a supervised policy. Based on the cross-domain knowledge base, we artificially construct a question-answering dataset that we hope the model generates aligned answers to SFT the pre-trained MLLM. For questions we want the model to answer well, we collect the answers we want the model to output to increase the probability of the model generating the expected answers. We choose currently popular and competitive open-source MLLMs for model construction and fine-tuning, including Qwen-VL (Qwen-VL-Chat), InternVL (InternVL2-8B and InternVL2-2B), Yi-VL (Yi-VL-6B), Deepseek-VL (DeepSeek-VL-7B-chat and DeepSeek-VL-1.3B-chat) and GLM-4 (GLM-4V-9B), all of which support English and Chinese. It cannot be denied that we mainly provide a method to build an MLLM suitable for wheat breeding, rather than having to choose a specific open-source MLLM, as models with better performance are constantly emerging. We used Scalable lightWeight Infrastructure for Fine-Tuning (SWIFT) to fine-tune the MLLM (The

ModelScope Team, 2024). We refer to the SWIFT recommended training script, in which the LoRA (Low-Rank Adaptation) fine-tuning method is selected, the number of epochs is set to 5, and no further hyperparameter adjustment is performed. All fine-tuning and inference were performed on four V100 NVIDIA GPUs with 32GB. Specifically, dataset D consists of the model input prompt and the answer that the breeder expects the model to output, denoted by $x$ and $y$ respectively. The model is denoted by $\pi_\theta$, and the length of $y$ is $T$. When prompt $x$ is given, the probability that the model generates answer $y$ can be expressed as formula (1). Where $\pi_\theta(y_t|x, y_{1:t-1})$ denotes the probability of the model outputting the $t$-th token given the input before the $t$-th token.

$$\pi_\theta(y|x) = \left[\prod_{t=1}^{T} \pi_\theta(y_t|x, y_{1:t-1})\right] \qquad (1)$$

The SFT stage uses the following loss to train the model. For simplicity, we call the model trained in this stage the SFT model.

$$L_{\text{SFT}} = -E_{(x,y)\sim D}[\log\pi_\theta(y|x)] = -E_{(x,y)\sim D}\left[\sum_{t=1}^{T} \log \pi_\theta(y_t|x, y_{1:t-1})\right] \qquad (2)$$

The second stage collects comparative data consisting of questions and different answers. Labels indicate their preferred answer given the input. A reward model (RM) is then trained to predict the breeder's preferred answer. Specifically, the model takes prompt and answer as input and outputs a scalar value. We use $r_\phi$ to denote the RM, and $r_\phi(x,y)$ denotes the scalar output of the RM given prompt $x$ and answer $y$. The dataset D used to train the RM consists of the model input prompt, the answer that the breeder wants the model to output, and the answer that the breeder does not want the model to output, denotes by $x$, $y_w$, and $y_t$ respectively. We are given data ($x$, $y_w$, $y_t$), and use the maximum likelihood estimation loss to train the RM. $\sigma$ denotes the sigmoid function. During training, K answers are selected each time and combined in pairs.

$$L_R(r_\phi) = -E_{(x,y_w,y_t)\sim D}[\log\sigma(r_\phi(x, y_w) - r_\phi(x, y_t))] \qquad (3)$$

In the third stage, proximal policy optimization (PPO) is used to optimize the policy for the RM. We use the output of the RM as a scalar reward and use the PPO algorithm to fine-tune the SFT model to optimize this reward (Schulman et al., 2017). Among them, the second and third stages can be carried out iteratively. We collect more comparison data on the current best policy, use it to train a new RM, and then train a new policy. In practice, most of our comparison data comes from supervised policies, and some of it comes from PPO policies. After training the RM, we have obtained an approximate reward function. Consider the model as a policy in the reinforcement learning (RL) problem, and the input and output of the model can be regarded as the state and action of RL respectively. We can use RL algorithms to train the model and train the model to output the answer with the highest reward. However, the RL training process is very unstable, so the SFT model is used as a reference model, denoted by $\pi_{\text{ref}}$. The KL divergence regularization term with the reference model is added to the objective function (4) to limit the update range of the model. Among them, the hyperparameter $\beta$ controls the degree of deviation from the reference model.

$$J_{r_\phi}(\pi_\theta) = E_{x\sim D,\, y\sim \pi_\theta}[r_\phi(x,y)] - \beta D_{KL}\big(\pi_\theta(y|x)\big|\pi_{\text{ref}}(y|x)\big) \qquad (4)$$

According to the objective function (4), the KL divergence term is expanded, and the final comprehensive reward is:

$$r(x,y) = r_\phi(x,y) - \beta(\log\pi_\theta(y|x) - \log\pi_{\text{ref}}(y|x)) \qquad (5)$$

The objective function is:

$$J_{r_\phi}(\pi_\theta) = E_{x\sim D,\, y\sim \pi_\theta}[r(x,y)] \qquad (6)$$

We created three different datasets: (1) The SFT dataset (75k) is created using multi-source datasets from the field, which contains questions and answers for training the SFT model. 80% of this dataset is used for training, and 20% is used for accuracy testing of wheat breeding model evaluation benchmark (phenotyping estimation task and environmental stress assessment task). (2) The RM dataset, with breeder rankings of model outputs, is used to train the RM. (3) The PPO dataset, without any breeder labels, which are used as inputs for RLHF fine-tuning. External domain knowledge base is mainly used by RAG methods. The dataset format refers to SWIFT (Table S4).

*3.2. WBLM evaluation benchmark*

In order to comprehensively evaluate the performance of MLLM in scientific breeding work, it is necessary to build a standardized machine and human evaluation benchmark. This benchmark aims to integrate challenges from various wheat breeding tasks. This benchmark aims to integrate challenges from various wheat breeding tasks. We have specifically designed many professional Chinese and English questions on wheat breeding and corresponding standard answers, covering five tasks: phenotyping estimation, environmental stress assessment, target germplasm screening, cultivation technique recommendation, and seed price query. The phenotyping estimation includes seven subtasks (Yield, SPAD, LAI, CH, CV, WH and PL), the environmental stress assessment includes two subtasks (WL and FVC), the target germplasm screening includes five subtasks: high quality (HQ), disease resistance (DS), drought resistance (DR), maturity period (MP), adapt to mechanized (AM), the cultivation technique recommendation includes two subtasks: cultivation technique (CT) and plant protection technique (PPT), and the seed price query includes one subtask seed price (SP) query. Through objective evaluation indicators and manual scoring and sorting, the evaluation team conducted a detailed evaluation of the answers of multiple MLLMs including WBLM, covering three aspects: accuracy, stability, and reasoning.

(1) Accuracy

We evaluate different subtasks under five tasks, respectively. For the phenotyping estimation task and the environmental stress assessment task, the RMSE and $R^2$ evaluation indicators are selected for regression tasks (estimating the values of FVC, SPAD, LAI, CH, CV, WH, and Yield), and the accuracy evaluation indicator is selected for the classification task (estimating the categories of WL and PL). For each subtask under the target germplasm screening task and cultivation technique recommendation task, each MLLM is tested multiple times (1k), and each test is manually judged whether the answer is correct. Then the proportion of the number of correct answers to the total number of tests is used as the evaluation indicator. For the seed price query task, we counted whether each result after multiple (1k) tests was consistent with the data in the knowledge base (price ± 10%), and took the proportion of the total number of consistent data to the total number of tests as the evaluation indicator.

(2) Stability

We evaluate the consistency and robustness of the answers. The evaluation indicator of stability is the proportion of times that satisfy the stability requirements in the total number of tests (1k), where the stability requirement is that the deviation of the numerical value in the answer is within ±10%, and whether the text satisfies the requirement is manually judged. Consistency: Since cross-domain data are related, the answers given after inputting different data from cross-domain data into the model should remain basically consistent. Robustness: The model's answer or performance should not change significantly when faced with small changes in input data. Robustness requires that the model should have a certain degree of fault tolerance.

(3) Reasoning

We evaluate the logical deduction, inductive reasoning, and explanation of the answers. The evaluation indicator of reasoning is the proportion of the total reasoning score of a single model after multiple tests (1k) to all scores. After one test, $x$ MLLM models output $x$ answers. After manually analyzing $x$ answers, we select a score from 1 to $x$ points for each answer and the score can only be selected once until all answers are scored. We sum up the answer scores of different models after multiple tests as the total reasoning score of the corresponding model. Then calculate the proportion of the total reasoning score of different models to all scores $((1 + x) * x / 2)$. Logical deduction: The model can deduce new conclusions or answers through logical reasoning based on known facts, rules and conditions. This requires the model to understand the logical relationships in the question and accurately apply these relationships to deduce answers. Inductive reasoning: The model can summarize general rules or patterns from a series of specific examples or observations and give reasonable answers based on them. Inductive reasoning requires the model to have the ability to abstract and generalize information. Explanation: The reasoning process is explained or visualized to a certain extent to help user understand the source and basis of the answer.

**4. Results and Discussion**

*4.1. Evaluation of WBLM*

The newly constructed wheat breeding model evaluation benchmark was used to evaluate closed-source MLLMs (Qwen 2.5, ERNIE Bot 4 Turbo, GPT-4o Plus and Gemini Pro 1.5, abbreviated as Qwen, ERNIE Bot, ChatGPT, Gemini) and WBLMs built based on different open-source MLLMs (Qwen-VL, InternVL, Yi-VL, Deepseek-VL and GLM-4) (Fig. 3-5, Table S5).

Accuracy evaluation: On various tasks of the evaluation benchmark, the comprehensive performance of WBLM based on InternVL2-8B is better than that of WBLM built on other open-source MLLMs, and is significantly better than closed-source ChatGPT, Gemini, and Qwen. Among them, for LAI evaluation, the $R^2$ based on InternVL2-8B is 0.795 and the RMSE is 1.302. The $R^2$ and RMSE of other open-source MLLMs range from [0.674, 0.772] and [1.542, 2.486], and the $R^2$ and RMSE of closed-source MLLMs range from [0.118, 0.142] and [3.61, 3.985], respectively. ChatGPT shows leading performance among open-source models, although still far behind the WBLM based on InternVL2-8B. In addition, we find that consistent with existing study (Z. Chen et al., 2024), the performance of MLLMs with small parameters in the same series is lower than that of MLLMs with larger parameters, such as InternVL2-8B and InternVL2-2B. We attribute this modest decline to the smaller size of the MLLMs.

Stability evaluation: Compared with closed-source MLLMs, WBLMs based on open-source models outperform in stability tests. Specifically, the WBLM based on InternVL2-8B achieved the best performance in consistency, with a stability score of 0.811, while the stability score of the lowest-performing Qwen 2.5 was only 0.053. The robustness scores of closed-sources MLLM ([0.829, 0.895]) are higher than those of WBLM based on open-source MLLM ([0.735, 0.802]), because closed-source MLLM cannot accurately answer relevant breeding questions in most tasks, and the answers are still wrong and similar when faced with slight changes in input data. However, among multiple WBLMs based on open-source MLLM, WBLM based on InternVL2-8B has a higher robustness score.

Reasoning evaluation: The WBLM based on InternVL2-8B outperforms other MLLMs in logical deduction and explanation evaluation, with the highest scores of 0.158 and 0.15, respectively. The WBLM based on GLM-4V-9B-chat outperforms other MLLMs in inductive reasoning evaluation, with the highest score of 0.16. Compared with accuracy and stability evaluation, reasoning evaluation tests the model's breeding decision support and problem-solving capabilities more. Excellent reasoning ability can enhance users' trust in MLLM and improve user satisfaction (J. Li et al., 2024).

InternVL2-8B based WBLM shows leading performance in most benchmark tests (Fig. 3-5, Table S5). However, the closed-source commercial MLLMs used in this study encountered significant difficulties in the wheat breeding evaluation benchmark test and performed poorly. The main reason is that closed-source commercial MLLM lacks domain knowledge related to wheat breeding, while open-source MLLM obtains it using a combination of domain knowledge technologies (Ming and Li, 2024). Although closed-source MLLMs represented by ChatGPT achieved better performance on the popular MLLM evaluation benchmarks (Fu et al., 2024; Y. Liu et al., 2024; Yue et al., 2024), we did not build WBLM based on closed-source MLLMs because these MLLMs do not support further implementation of the process of this study. If closed-source MLLMs support richer fine-tuning and downstream task customization, they may gain greater commercial value and more users. Although all open-source MLLMs can handle multimodal tasks, different network architectures have different abilities to solve semantic alignment between modalities when integrating visual and textual data, thus affecting the accuracy of the final results. In addition, different training processes, tricks, and private training data are also important factors affecting model performance (D. Huang et al., 2024). In the future, researchers can build breeding models based on MLLM with better performance to provide stronger breeding assistance efficiency and results.

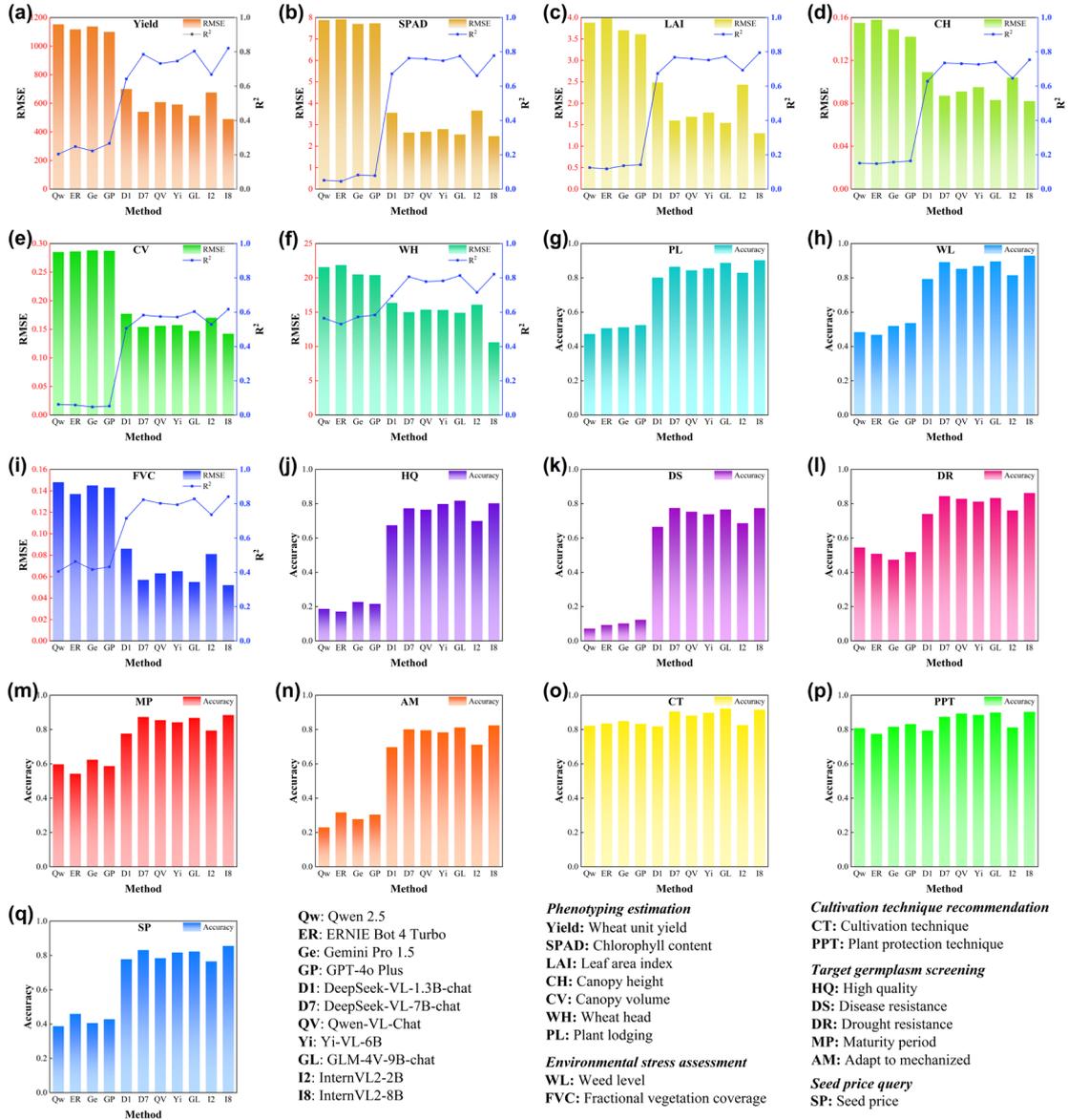

Fig. 3. Comparison of different MLLMs on the evaluation benchmark (accuracy).

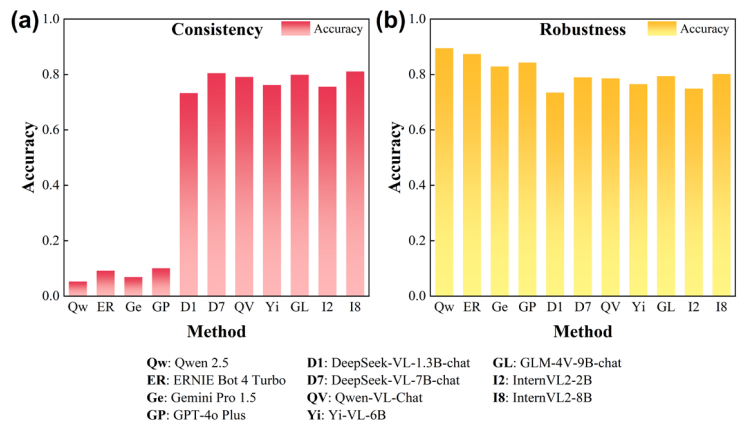

Fig. 4. Comparison of different MLLMs on the evaluation benchmark (stability),

which includes consistency and robustness.

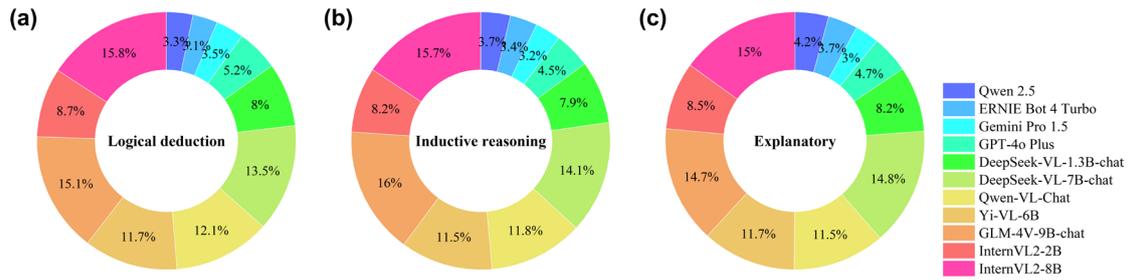

Fig. 5. Comparison of different MLLMs on the evaluation benchmark (reasoning). This figure shows the total reasoning score as a proportion of all scores after multiple tests of a single MLLM.

*4.2. Impact of domain knowledge on breeding goals*

*4.2.1. Contribution of the combination of domain knowledge and technology to breeding goals*

We evaluated WBLM on the wheat breeding model evaluation benchmark and observed strong performance of WBLM based on InternVL2-8B. In this section, we conduct ablation experiments to test the performance of different domain knowledge technologies combinations in different tasks (Fig. 6 and Table S6).

For the phenotyping estimation task, Yield, SPAD, LAI, CH, and CV show poor performance under the MLLM-only method due to the same lack of domain knowledge. Compared with the MLLM-only method, the MLLM and SFT method, the MLLM and RAG method, and the MLLM and RLHF method all greatly improve the prediction performance of these subtasks. Additionally, for the WH and PL subtasks, MLLM has certain detection and classification capabilities. With the addition of SFT, RAG, and RLHF technologies, the performance is further improved to achieve relatively high $R^2$ and low RMSE. For the environmental stress assessment task, the results showed that accurate classification of WL was simpler than specific values estimation of FVC. When the method combining MLLM, SFT, RAG, and RLHF was used, the best performance for WL prediction was 0.929 (accuracy), and the best performance for FVC prediction was 0.841 ($R^2$) and 0.052 (RMSE).

When only MLLM is used, the prediction results for the target germplasm screening task and the seed price query task are poor. The main reason is that the MLLM lacks domain knowledge and price dynamic data, so it cannot answer accurately. Furthermore, the results show that when the MLLM is combined with one or more domain knowledge technologies (SFT, RAG, RLHF), better prediction performance can be achieved. For the cultivation technique recommendation task, because the MLLM has been trained with many images and texts from various data sources, and some wheat cultivation technique knowledge is also covered, thus it can answer some questions well. It is undeniable that when one or more domain knowledge technologies (SFT, RAG, RLHF) are combined, the cultivation technique recommendation is more accurate and the data is traceable.

The method combining MLLM, SFT, RAG and RLHF performs better in different subtasks of the five tasks. It is worth mentioning that by analyzing the prediction results of the above tasks, it can be clearly found that hallucination exists when only the MLLM method is used, which directly reflects the necessity of combining MLLM with domain knowledge technology. With the increase of domain knowledge technology, the $R^2$ of the methods in the phenotyping estimation and environmental stress assessment tasks gradually increased, and the RMSE decreased. Simultaneously, the consistency of the prediction results in the target germplasm screening, cultivation technique recommendation, and seed price query tasks gradually increased. This shows that all methods based on domain knowledge technology can enhance the model's understanding of specific fields to some extent, improve data interpretation capabilities, and thus optimize prediction accuracy and reliability.

However, when the task involves many rare or long-tail scenarios (e.g., phenotyping estimation tasks), the method combining MLLM, SFT and RLHF outperforms the method combining MLLM, RLHF and RAG and the method combining MLLM, SFT and RAG. When the task involves obtaining knowledge from many structured and domain-related data (e.g., the seed price query task), the method combining MLLM, RLHF and RAG and the method combining MLLM, SFT and RAG outperform the method combining MLLM, SFT and RLHF. This may be due to the unique advantages of different domain knowledge technologies in dealing with specific tasks. This result is consistent with previous studies (Balaguer et al., 2024; Giuffre et al., 2024; J. Li et al., 2024), which showed that SFT directly optimizes model output by annotating data, RAG integrates external multimodal knowledge to improve answer quality, and RLHF iteratively optimizes generated content to improve task orientation and adaptability. Therefore, the method combining MLLM, SFT, RAG, and RLHF combines the advantages of different technologies to achieve better performance. It should be noted that this method may be affected by the update frequency and accuracy of the knowledge base, resulting in a decrease in the quality of answers (Siriwardhana et al., 2023). In addition, RLHF requires a large amount of interaction data to optimize the MLLM, and sample efficiency issues may occur in the process (Yuan et al., 2024). Future study can integrate more advanced domain knowledge technologies (e.g., knowledge graphs) to extract better domain knowledge to achieve more accurate generation prediction capabilities for different breeding tasks.

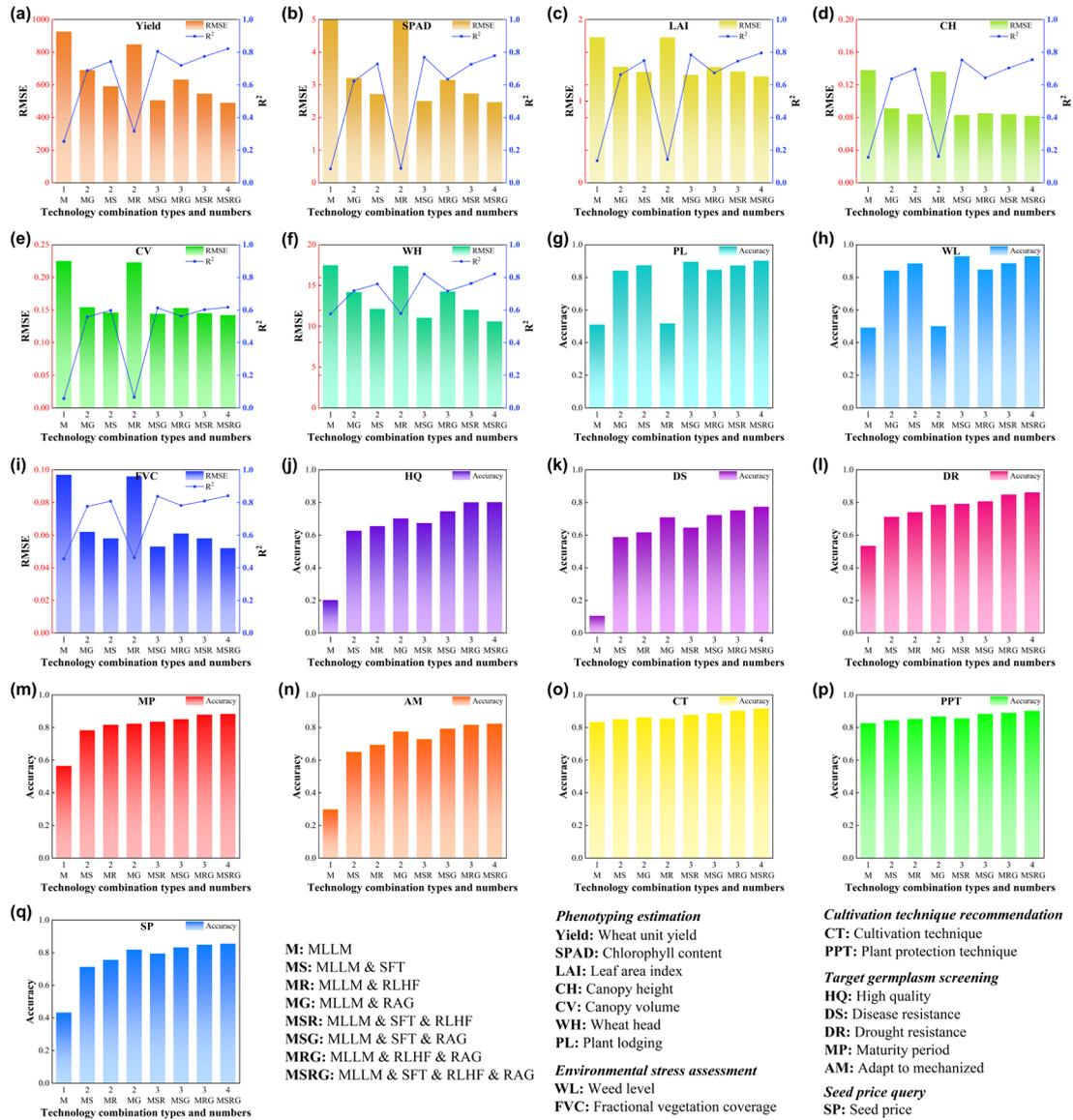

Fig. 6. Prediction performance of different tasks for methods with different combinations of domain knowledge technologies.

*4.2.2. Performance of cross-domain data combination in wheat yield prediction*

When only RS data is used, the $R^2$ of wheat yield prediction is 0.707 and the RMSE is 599.532 kg/ha (Fig. 7 and Table S7). The RS data comes from processed data from different UAV sensors. As shown in many previous studies, different information such as VI (spectral), FVC (environment), CH (canopy structure) have become the most commonly used RS indicators in crop yield prediction due to their stable and superior performance (Roth et al., 2022; Skobalski et al., 2024; T. Xu et al., 2023). However, a wide range of environmental conditions such as growing soil, water availability and atmospheric conditions, as well as different germplasm genes of wheat, also have a significant impact on yield, which may reduce model performance.

The combination of RS and phenotyping data can significantly improve the prediction accuracy compared with using RS data alone. The canopy structure features (LAI) and physiology and biochemistry features (SPAD) obtained from phenotyping data contain independent information about canopy growth and structure, as well as information that indirectly reflects the relative content of current chlorophyll in plant leaves. Furthermore, phenotyping data have the advantage of higher accuracy to a certain extent, and can directly capture more subtle changes and differences, which is particularly important for accurately assessing wheat growth conditions and predicting yields. Therefore, the combination of RS and phenotyping data can improve the prediction accuracy. Many previous studies have verified the potential of coupled RS and phenotyping information in crop yield prediction (Duan et al., 2017; Jin et al., 2024). The combination of RS and weather data slightly improved the prediction accuracy compared with RS data alone, with an increase of 0.014 in $R^2$ and a decrease of 9.84 kg/ha in RMSE. The information from weather data tends to add supplementary information to RS information in wheat yield prediction, but to a lesser extent than the information contained in phenotyping data. It shows that RS information and weather information at different growth stages are prone to information overlap, further affecting the prediction ability. When the relevant information based on germplasm data (high yield, HQ, DS, DS, MP and AM) was added to the information from RS data, the prediction accuracy of all methods was slightly improved. It shows that using macro-growth conditions provided by RS data and micro-genetic traits provided by germplasm data can provide a more comprehensive assessment of crop growth potential and yield.

The yield prediction performance from the combination of RS, phenotyping, and weather data; RS, phenotyping, and germplasm data; and RS, weather, and germplasm data is better than the combination of the two data types, with $R^2$ varying from 0.734 to 0.808 and RMSE ranging from 494.967 kg/ha to 572.755 kg/ha. These results indicate that RS, phenotyping, weather, and germplasm information provide unique and complementary information that is helpful for wheat yield prediction. These results at least partially further explain the benefits of using cross-domain data in wheat yield prediction. For this benefit, we further utilized all the data obtained (RS, phenotyping, weather, germplasm), and the results showed that the yield prediction performance was the best, with an $R^2$ of 0.821 and an RMSE of 489.254 kg/ha. Related studies have also shown that using one or more data from RS, environment and germplasm can improve the performance of model yield prediction (Feng et al., 2024; Maimaitijiang et al., 2020; Tian et al., 2022). These studies use data fusion as one of the data-level solutions to improve the accuracy of yield prediction. It is undeniable that as the data scale increases, the model can usually achieve improved performance, but this increases the complexity of the model, requiring more powerful computing resources and more complex algorithms (J. Li et al., 2024). And data quality and missing issues directly affect the prediction accuracy of the model (Yang et al., 2023). In addition, obtaining high-quality yield data requires a lot of field experiments and data processing, which increases the cost of data acquisition (Ruan et al., 2022). Although data fusion can improve the generalization ability of the model, the model may still experience performance degradation when facing a completely new planting environment or germplasm (Gu et al., 2024).

The predicted values of wheat yield obtained by the methods (trained using data from different data sources) were compared with the corresponding observed values using the scatter plot (Fig. 7). As expected from the comparable $R^2$ and RMSE values, the values obtained by the method trained only with RS data are widely scattered on both sides of the regression line, while the values obtained by the method trained with all multi-source data are closely around the regression line. It is worth noting that the data from the China Rural Statistical Yearbook show that the yield per unit area of wheat in Zhejiang Province in 2022 is 4230.2 kg/ha (National Bureau of Statistics of China, 2022), where the points to the right of the purple dashed line in Fig. 7 represent germplasms that exceed the yield per unit area. The germplasms with yields exceeding 4230.2 kg ha$^{-1}$ are from China, Japan, Argentina, Italy, Turkey, and Australia. These germplasms showed higher FVC, CH, CV, and WH compared to other germplasms. It should be emphasized that these germplasms achieved higher yields in an environment with stronger weed competition, thus indicating that these germplasms have stronger competitiveness and adaptability. These wheat germplasms are more valuable in production because their growth shows a certain degree of freedom from weed control activities, reducing production costs and improving yield sustainability.

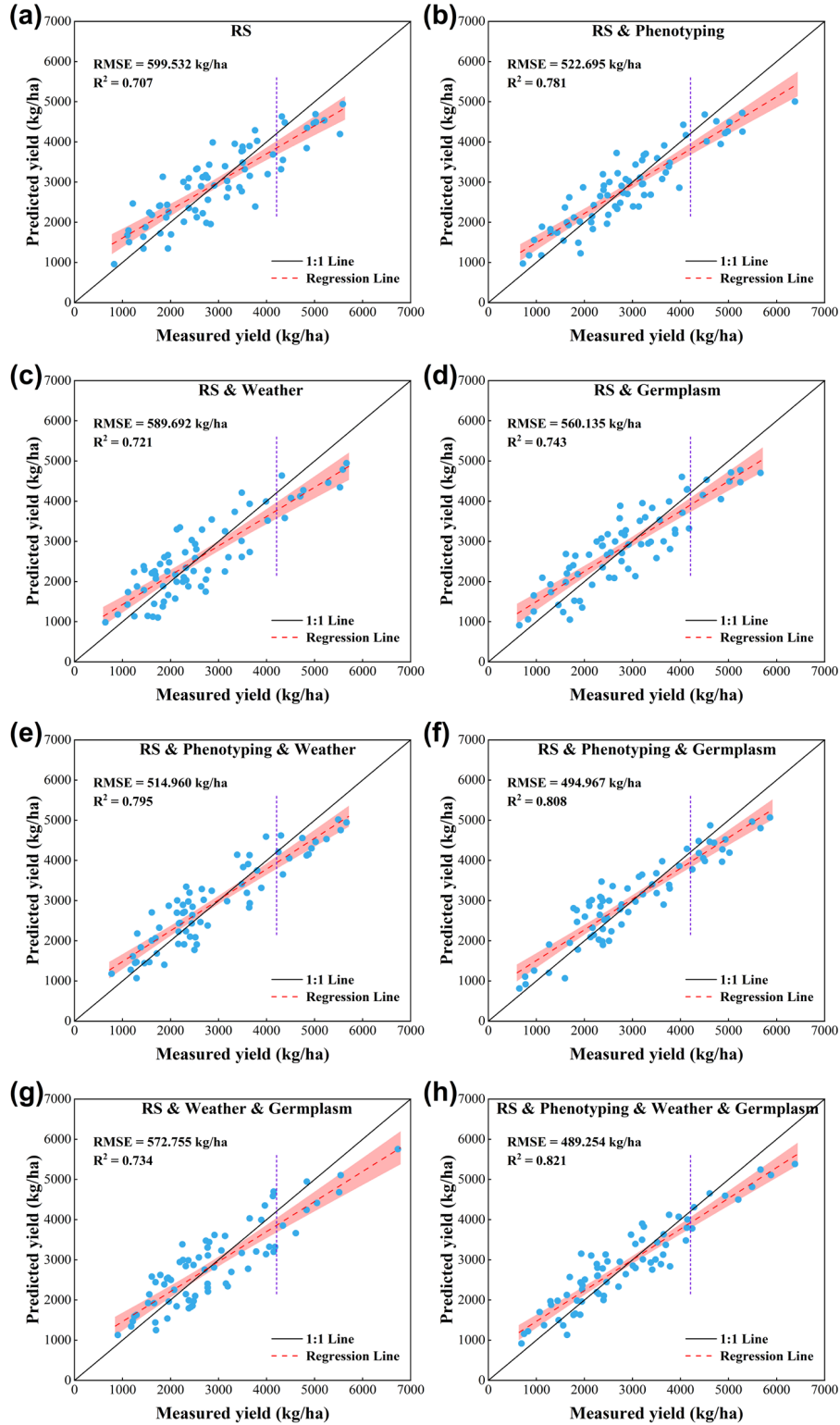

Fig. 7. Cross-validation scatter plot of measured and predicted wheat yield. The black solid line indicates a 1:1 relationship. The blue points to the right of the purple dashed line represent germplasms with wheat yields exceeding 4230.2 kg ha$^{-1}$.

*4.3. Breeding goals response to WBLM*

We compare WBLM with the easily accessible MLLM: Qwen 2.5, ERNIE Bot 4 Turbo, GPT-4o Plus, and Gemini Pro 1.5 in different scenarios (the five tasks of this study) to explore wheat breeding goals response capabilities of MLLM. Since our WBLM and all other MLLMs support both Chinese and English, we chose Chinese and English for testing. The results show that WBLM, Qwen and ChatGPT can correctly and automatically adjust to the corresponding language to answer according to the user's query language, while the other two MLLMs need to explicitly specify the answer language. We aim to demonstrate the practicality and versatility of WBLM and other MLLMs in real breeding applications, providing analysis and insights from the perspective of actual user experience whenever possible.

We can use WBLM and commercial MLLMs to estimate the yield of different germplasms (Fig. 8). Specifically, we provide different phenotyping images and related growth information. Except for WBLM, other MLLMs cannot answer the questions accurately, and WBLM's answers are simple and direct. Some MLLMs cannot answer correctly, but ideas and methods are provided to further answer the question. The results show that WBLM has unique and superior capabilities in yield phenotyping estimation than other MLLMs.

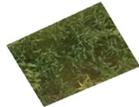
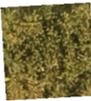

Fig. 8. Examples of phenotyping (yield) estimation on different MLLMs.

We can use WBLM and commercial MLLMs to screen target germplasm. We test five target germplasm characteristics: HQ, DS, DR, MP, and AM, either individually or in combination. Regarding the screening of AM (plant height) (Fig. 9a), Qwen and ERNIE Bot were unable to answer correctly, while other MLLMs were able to answer correctly. For the HQ screening (Fig. 9b), Qwen and ERNIE Bot cannot correctly answer the specific variety name and preservation organization, while other MLLMs can answer correctly. For the MP and AM screening (Fig. 9c), we found that except for Gemini, other MLLMs can correctly output specific wheat variety names that satisfy the requirements. For DS, DR and HQ screening (Fig. 9d), each MLLM can correctly output specific wheat varieties. Analyzing the above results, except for WBLM, which has a clear data source, the data sources of other MLLMs are unknown. In addition, the answers of different MLLMs vary greatly, so users need to verify and refer to them with caution. In addition, if MLLM cannot answer the question correctly by self-judgment, some solutions or methods will be provided.

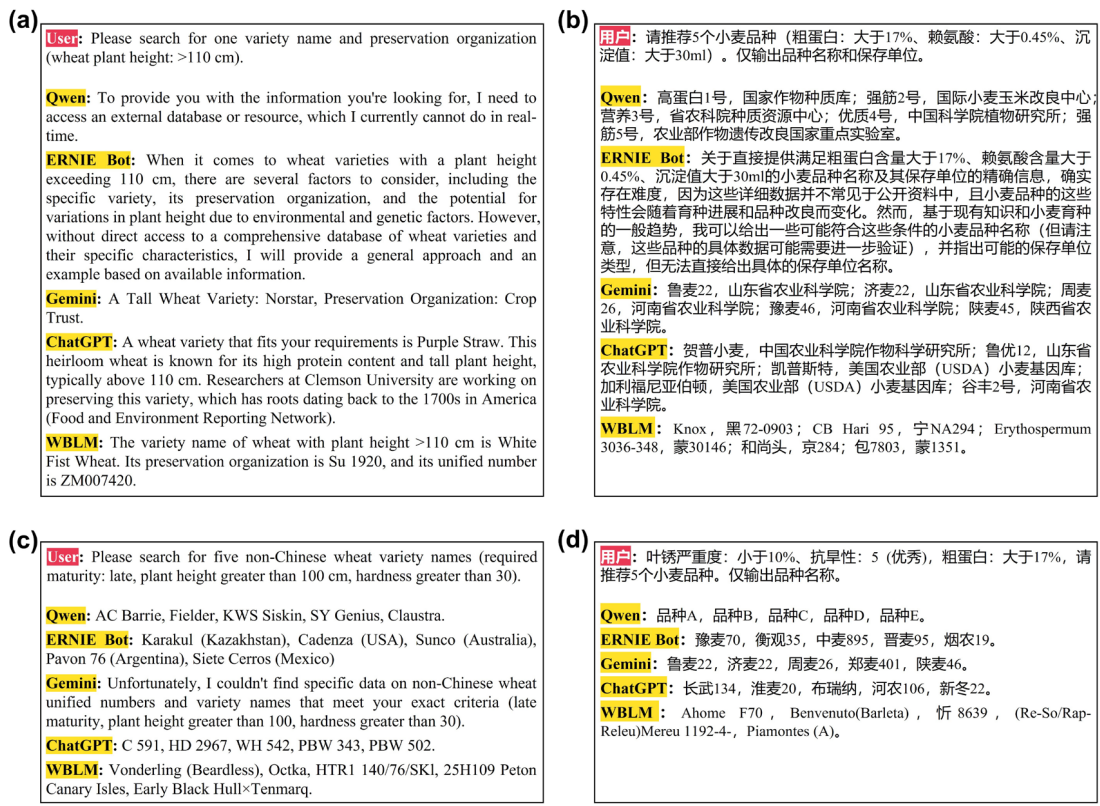

Fig. 9. Examples of target germplasm screening on different MLLMs.

We can use WBLM and commercial MLLMs to assess the environmental stress

(FVC and WL) of different germplasms (Fig. 10). Only WBLM can accurately output the value of FVC and the level of WL. ChatGPT can answer questions in the environmental stress assessment, but the answers are inaccurate. ERNIE Bot and Gemini can only partially answer and the answers are inaccurate. Qwen cannot answer either. Furthermore, we found that when there was no word limit on the answer, WBLM had a deeper understanding of the knowledge in the field of wheat breeding and the answers were more focused and specific.

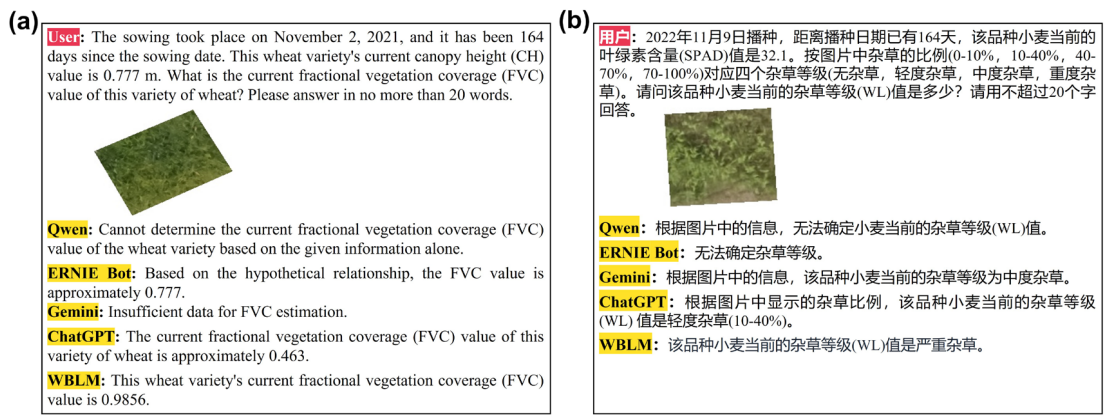

Fig. 10. Examples of environmental stress assessment on different MLLMs.

We can use WBLM and commercial MLLMs to recommend efficient, innovative and sustainable PT and PP techniques (Fig. 11). Focus on their ability to accurately answer questions under different techniques, different growth stages, and different growth environments. The scope of the recommendation: From soil preparation technique before wheat planting, to wheat sowing time, sowing amount and depth; from wheat fertilization management (base fertilizer), to when to apply topdressing and what fertilizer to use; from how to effectively control weeds in field management, to wheat disease and pest control in specific growth periods; from introducing wheat soil testing and formulation technique, to introducing wheat "one spray and three preventions" technique. The results show that the performance of WBLM is on par with Qwen and ChatGPT, and slightly better than the answers of other MLLMs. This reflects WBLM's superior understanding and reasoning of PT and PP technique, resulting in better recommendation results than the answers of other MLLMs.

(a) **User:** Please recommend soil preparation techniques for growing wheat. Please answer in 100 words or less.

**Qwen:** For wheat, prepare the soil by first removing weeds and debris. Incorporate organic matter like compost to improve soil structure. Plow deeply to break up compacted layers and then disc to create a fine seedbed. Adjust pH to around 6.0-7.5 using lime or sulfur as needed. Test soil fertility and apply balanced fertilizers according to recommendations. Ensure good drainage to prevent waterlogging.

**ERNIE Bot:** For growing wheat, begin by testing your soil's pH and nutrient levels to determine what amendments are needed. Ideally, wheat prefers a slightly acidic to neutral soil with a pH between 6.0 and 7.0. Incorporate organic matter like compost or manure to improve soil structure and fertility. Till the soil to a depth of 6-8 inches to break up clumps and loosen the soil. Make sure to remove any weeds, rocks, or debris that may hinder growth. Finally, level the soil to ensure even drainage and a smooth seedbed for planting.

**Gemini:** Soil Preparation for Wheat: Tillage: Plow the soil deeply to break up clods and aerate it. Fertilization: Apply balanced fertilizer based on soil testing. Liming: Adjust soil pH if necessary using lime. Weed Control: Use herbicides or mechanical methods to remove weeds before planting. Moisture Management: Ensure proper soil moisture levels for germination. Residue Management: Incorporate crop residues to improve soil health. For more specific recommendations, consult with local agricultural experts or extension services.

**ChatGPT:** For optimal wheat growth, start by plowing the soil to a depth of 15-20 cm to enhance aeration and root development. After plowing, use harrowing to break clods and level the soil. Incorporate organic matter, such as compost or manure, to improve soil fertility. Soil testing is essential to adjust pH and nutrient levels. Apply balanced fertilizers based on the results. Maintain good drainage and ensure fine tilth for proper seedbed preparation. Timely soil preparation supports uniform seed germination and healthy growth.

**WBLM:** Perform regular soil testing to identify nutrient deficiencies and pH imbalances. Use lime to raise pH of acidic soils and sulfur to neutralize excess sulfur compounds. Add organic matter to improve soil structure and water retention capacity. Remove debris and rocks to ensure proper root growth. Cover soil with mulch to retain moisture and suppress weeds.

(b) **用户：** 请推荐小麦播种时间，播种量和深度。请在100字以内回答。

**Qwen：** 小麦适宜播种时间为10月中旬至11月初，每亩播种量约15-20公斤，播种深度2-3厘米。

**ERNIE Bot：** 小麦播种时间一般在秋季的9月下旬到10月上旬，但具体需根据当地气候和品种确定。播种量每亩约为20-25斤，土壤肥力和墒情会影响播种量。播种深度一般为3～5厘米左右，根据地力和墒情适当调整。请注意，播种时间和量需结合当地实际情况和专家建议确定。

**Gemini：** 播种时间：控制在当地造墒期前，确保小麦有足够的时间根系生长。播种量：根据品种特性和地力条件而定，一般高产地块应确保40万/亩左右，用种量10-15公斤左右。播种深度：2-3厘米，过深过浅均不利于发芽。

**ChatGPT：** 小麦的播种时间一般在秋季，具体时间因地而异，北方地区通常在10月上中旬，南方则在11月左右。播种量通常为每亩15-25公斤，视品种和土壤条件调整。播种深度一般为3-5厘米，土壤墒重或干旱时可适当加深到5-6厘米，以确保种子与土壤充分接触，促进均匀发芽和早期生长。

**WBLM：** 冬小麦在9月下旬至10月上旬播种，春小麦在3月下旬至4月上旬播种。播种量与深度应根据土壤肥力和品种特性来确定，一般每亩播种量在10-15公斤之间，播种深度控制在3-5厘米。

(c) **User:** Please introduce wheat fertilizer management (base fertilizer). Your answer should not exceed 100 words.

**Qwen:** For wheat base fertilizer, apply a balanced mix of N-P-K. Use about 50-70% of total nitrogen, full phosphorus, and potassium. Typically, apply 40-60 kg N/ha, 20-40 kg P2O5/ha, and 20-40 kg K2O/ha. Incorporate into the soil before planting. Adjust based on soil tests and crop needs.

**ERNIE Bot:** Wheat fertilizer management, especially base fertilizer application, is crucial for optimal crop growth and yield. Base fertilizer should be applied before planting to enrich the soil with essential nutrients like nitrogen, phosphorus, and potassium. This ensures that wheat plants have adequate nutrients throughout their growth cycle, supporting strong root development, increased disease resistance, and ultimately higher grain yields. The specific fertilizer blend and application rate should be based on soil test results and local agricultural recommendations.

**Gemini:** Base fertilizer is applied before planting and provides essential nutrients for the entire growing season. Key points for wheat: Soil Testing: Determine nutrient needs based on soil analysis. Nutrient Balance: Apply a balanced fertilizer containing N, P, and K. Placement: Broadcast or apply in bands near the seed. Rate: Adjust application rate based on soil fertility and desired yield. Timing: Apply fertilizer before planting or during the fall for winter wheat. Consideration: Avoid excessive nitrogen application to prevent lodging. Consult with local agronomists or extension services for specific recommendations based on your region and soil conditions.

**ChatGPT:** Base fertilizer for wheat is typically applied before planting to improve soil fertility. It usually includes a mix of nitrogen (N), phosphorus (P), and potassium (K). Phosphorus and potassium are applied in full, while a portion of nitrogen is included to support early growth. Organic manures can also be incorporated to boost soil structure and nutrient availability. Application rates vary based on soil tests, but generally, 60-80 kg/ha of nitrogen, 60-70 kg/ha of phosphorus, and 30-40 kg/ha of potassium are used. Proper base fertilization ensures strong root development and enhances yield potential.

**WBLM:** Before sowing, apply sufficient base fertilizer to provide the nutrients needed for wheat growth. Base fertilizer should be mainly organic fertilizer, supplemented by chemical fertilizer. Generally, 100 to 120 kilograms of 51% compound fertilizer is applied per mu.

(d) **用户：** 小麦种后一般什么时候追肥? 选用什么肥? 请在100字以内回答。

**Qwen：** 小麦种后追肥通常在返青至拔节期进行，主要使用尿素，也可在孕穗期和灌浆期适量喷施叶面肥如磷酸二氢钾和芸苔素。

**ERNIE Bot：** 小麦种后，通常在拔节期（分蘖高峰后）和孕穗期进行追肥。拔节期以氮肥为主，如尿素，促进茎秆生长；孕穗期补充磷酸肥，如复合肥，增强籽粒灌浆。具体时间和用量需依据土壤测试和植株生长情况调整。

**Gemini：** 小麦种后一般在 返青期 和 拔节期 进行追肥。返青期追肥：主要以氮肥为主，如尿素、碳酸氢铵等，促进分蘖，提高成穗率。拔节期追肥：仍以氮肥为主，配合磷钾肥，促进穗粒形成，增加粒重。

**ChatGPT：** 小麦播种后一般在返青期和拔节期进行追肥。返青期每亩施氮肥，以促进分蘖和植株早期生长，通常每亩施用10-15公斤尿素。拔节期是关键的营养生长期，追施氮磷钾复合肥，促进茎叶和籽粒发育，每亩施用15-20公斤尿素或复合肥。根据土壤和生长状况调整施肥量，有助于提高产量和品质。

**WBLM：** 一般在年后返青时到拔节时追肥，选用速效肥，如尿素、硝基水溶肥、碳酸氢铵等高效氮肥。

(e) **User:** Please recommend how to effectively control weeds in wheat field management. Please answer in 100 words or less.

**Qwen:** Employ integrated weed management: use pre-and post-emergence herbicides, mechanical control like hoeing, and crop rotation. Select competitive wheat varieties, plant densely, and scout fields regularly. Consider biological controls and keep farm equipment free of weed seeds. Integrated Weed Management ensures sustainable, chemical-reduced weed control.

**ERNIE Bot:** To manage weeds in wheat fields, adopt a holistic approach: start with clean seed, use crop rotation to disrupt weed cycles, maintain proper soil tillage to bury weed seeds, apply mulch to suppress germination, and selectively use herbicides targeting specific weed species. Early detection and control are crucial. Follow label instructions for herbicide use, minimizing environmental impact. Regular monitoring and adjusting management strategies based on weed pressure will ensure optimal wheat growth and yield.

**Gemini:** Fight wheat weeds with a multi-pronged attack! Use clean seed to avoid weed introduction. Combine methods like crop rotation, pre-emergent herbicides to stop early growth, and post-emergent options for specific weeds. Consider tillage before planting, but time it right to avoid harming wheat. This integrated approach keeps your wheat field weed-free!

**ChatGPT:** To control weeds in wheat fields, use pre-emergent and post-emergent herbicides strategically. Rotate crops to disrupt weed cycles and prevent resistance. Implement mechanical weeding through cultivation and hoeing. Plant cover crops to suppress weed growth and maintain optimal wheat planting density to limit weed space. Regularly monitor and manually remove weeds as needed.

**WBLM:** Choose the right pesticide and carry out pre-winter chemical control in time. Field management should adapt to local conditions and provide classified guidance. Combine control and promotion, control and promote, so that vigorous seedlings can be transformed into strong ones, strong seedlings can grow steadily, weak seedlings can be upgraded, and balanced growth of various wheat fields can be achieved. Finally, timely inter-cultivation and loosening of the soil should be carried out to break the compaction, increase the ground temperature, and promote weak to strong.

(f) **用户：** 请介绍小麦纹枯病拔节初期防治方法。请在100字以内回答。

**Qwen：** 拔节初期防治小麦纹枯病，应喷施噻呋酰胺悬浮剂等药剂，并确保药液覆盖基部茎。同时，可使用5%井冈霉素1000倍液喷雾。若病株率达10%，需立即采取措施。

**ERNIE Bot：** 小麦纹枯病拔节初期防治方法：农业防治：早施速效氮肥、磷钾肥，清沟理墒，降低田间湿度。药剂防治：选用井冈·蜡芽菌、苯甲·丙环唑等药剂，按建议浓度喷雾，发病重田块首次施药后10天再施一次，注意喷药质量。确保药液流淌到麦株基部，提高防治效果。

**Gemini：** 小麦纹枯病拔节初期防治方法：农业防治：清除病株残体，合理轮作，选用抗病品种。药剂防治：可用井冈霉素、三唑酮等系列药剂喷雾，重点喷施基部茎。生物防治：可结合使用枯草芽孢杆菌等生物菌剂。

**ChatGPT：** 小麦拔节初期防治纹枯病的方法包括：清除病残体，减少病原菌；合理密植，增强通风；加强水肥管理，避免氮肥过量，可喷施三唑类杀菌剂，如苯醚甲环唑，每隔7-10天喷1-2次，有效抑制病菌。

**WBLM：** 小麦纹枯病拔节期间病株率达10%时，可以采取主动施药防治的方法。常用的防治药剂有井冈·蜡芽菌、噻呋酰胺、戊唑醇、丙环唑、烯唑醇、井冈霉素、多抗霉素等。

(g) **User:** Please introduce the soil testing and formula fertilization technology. Please answer within 100 words.

**Qwen:** Soil testing involves analyzing soil samples to determine nutrient levels. Formula fertilization uses these results to prescribe specific nutrient ratios tailored to crop needs. Test annually; collect samples from 0-20 cm depth. Adjust pH if necessary. Apply nutrients based on test results and crop requirements. Monitor crop growth and adjust formulas as needed.

**ERNIE Bot:** Soil testing involves analyzing soil samples to determine nutrient levels and other properties. Based on the results, formula fertilization technology prescribes specific blends of fertilizers tailored to the soil's needs. This ensures optimal plant growth by providing essential nutrients in balanced amounts, avoiding over-fertilization and minimizing environmental impact. It promotes sustainable agriculture by enhancing soil health and crop productivity.

**Gemini:** Soil testing involves analyzing a soil sample to determine nutrient levels, pH, and other properties. Based on these results, a formula fertilizer is created to provide the exact nutrients needed for optimal plant growth. This targeted approach helps to minimize nutrient waste and environmental impact. Key steps: Soil Sampling: Collect soil samples from different areas of the field. Laboratory Analysis: Send samples to a lab for testing. Nutrient Recommendations: Receive a report detailing nutrient needs. Formula Fertilizer Creation: Develop a custom fertilizer blend. Application: Apply the fertilizer according to recommendations.

**ChatGPT:** Soil testing and formula fertilization technology involves analyzing soil samples to determine nutrient levels and soil pH. Based on the results, a customized fertilizer plan is created, ensuring balanced nutrient supply for crops. This approach helps avoid over- or under-application of nutrients like nitrogen, phosphorus, and potassium. It improves fertilizer use efficiency, promotes sustainable agricultural practices, and enhances crop yields while minimizing environmental impact. The technology is widely used to optimize crop nutrition and boost productivity by addressing specific soil deficiencies and crop needs.

**WBLM:** Soil testing and formula fertilization technology is based on wheat field tests and fertilizer field trials. The application amount, fertilization period and application method of fertilizers (mainly nitrogen, phosphorus and potassium fertilizers) are determined on the basis of reasonable application of organic fertilizers in wheat fields with different soil fertility levels.

(h) **用户：** 请介绍小麦一喷三防技术。请在100字以内回答。

**Qwen：** 小麦一喷三防技术是在抽穗扬花期喷施杀菌剂、杀虫剂和植物生长调节剂，以防治病虫害、干热风和增粒重。常用配方包括15%多菌灵5%吡虫啉、0.2%磷酸二氢钾。

**ERNIE Bot：** 小麦"一喷三防"技术是小麦生长中后期，通过一次性喷施杀虫剂、杀菌剂、植物生长调节剂（如叶面肥）等混配剂，达到防病、防虫、防干热风，实现增粒增重，确保小麦高产的关键技术措施。该技术由农业农村部门推广，是农业防灾减灾、稳产增产的重要措施。

**Gemini：** 在小麦抽穗扬花期，通过一次喷雾，达到防治病虫害、防干热风、促进灌浆增产的目的。具体做法：将杀虫剂、杀菌剂、叶面肥等混合喷施能有效防治小麦常见的病虫害（如锈病、白粉病、蚜虫），减轻干热风危害，促进籽粒饱满，提高产量。

**ChatGPT：** 小麦生长关键期通过一次喷药，防治病害、虫害并防干热风。通常在小麦抽穗扬花期喷施杀菌剂、防治病和植物生长调节剂，防治赤霉病、锈病、蚜虫等，增强抗逆性，防止干热风影响。这项技术能提高小麦产量和品质，减少病虫害及气候不利因素带来的损失。

**WBLM：** 小麦"一喷三防"技术是在小麦抽穗后至灌浆期，在叶面喷施杀菌剂、杀虫剂、植物生长调节剂形成混配液，通过一次施药实现防病、防虫、防早衰，进一步提高粒重。

Fig. 11. Examples of cultivation technique recommendation on different MLLMs.

We can use WBLM and commercial MLLMs to query the seed varieties, specifications and prices at a specified observation point and time (Fig. 12). It can be seen that except for WBLM, none of them can provide accurate answers. Other MLLMs choose to give factors that affect wheat seed prices and suggest methods to obtain relevant data. Through this experiment, we found that for highly dynamic information such as price data, MLLMs need to rely on external data sources to ensure the real-time and accuracy of the data. Although MLLMs can provide trend analysis, principle explanations, or forecasting methodologies based on historical data, they are unable to capture new market conditions, stock prices, commodity prices, or other information that requires real-time updates. Real-time and price data are usually collected and published by specialized financial data providers, market analysis companies or government statistics departments. These data often need to be obtained through subscription services, API interfaces or public market reports.

(a)

**User**: Address: Miyun District, Beijing, Time: 2024-06-01, Please answer the price (CNY/bag), variety, and specifications (CNY/bag) of wheat seeds?

**Qwen**: I don't have real-time data access for specific local market prices, including wheat seed prices in Miyun District, Beijing, on June 1, 2024. To get accurate information about wheat seed prices, varieties, and specifications, you can follow these steps: Local Agricultural Supply Stores, Agricultural Cooperatives, Online Platforms, Government Resources. If you need general information about common wheat seed varieties and typical prices as of my last update in 2023, I can provide that. However, please note that actual prices may vary significantly due to factors such as market demand, supply, and weather conditions. Would you like me to provide this general information instead?

**ERNIE Bot**: As an AI language model, I do not have access to real-time market prices or specific product information for individual businesses or regions. However, I can provide you with a general overview of wheat seed prices, varieties, and specifications in China, which may vary depending on factors such as market conditions, quality, and the specific variety.

**Gemini**: Agricultural product prices, including wheat seeds, can fluctuate based on various factors such as seasonality, supply and demand, and specific varieties. To obtain the most accurate and up-to-date information regarding the price, variety, and specifications of wheat seeds in Miyun District, Beijing, for June 2024, I recommend consulting the following resources: Local Agricultural Departments, Seed Retailers and Distributors, Online Agricultural Marketplaces. By utilizing these resources, you can gather the most reliable and relevant information regarding wheat seed prices, varieties, and specifications in Miyun District, Beijing, for June 2024.

**ChatGPT**: The price of wheat seeds can vary depending on the variety and specifications, as well as the supplier. For a specific location like Miyun District, Beijing, and a particular date, it's best to consult local agricultural suppliers or seed companies for the most accurate and up-to-date information.

**WBLM**: Price: 150 CNY/bag, Variety: Nongda 3486, Specification: 25 kg/bag.

(b)

**用户**：地点：四川成都市，时间：2024-07-24，请仅回答价格，品种，规格是（小麦种子）？

**Qwen**：对于具体的小麦种子的价格、品种和规格信息，这些数据通常会根据市场情况、供应商以及具体的品种而有所不同。在现实中，这类详细信息往往需要通过当地的农业合作社、种子销售商或者官方农业部门来获取。

**ERNIE Bot**：针对您提出的关于2024年7月24日在四川成都市小麦种子的价格、品种及规格的问题，由于市场价格受多种因素（如供需关系、季节变化、政策调整、种植技术更新等）影响，且这些信息是实时变动的，我无法直接提供确切的实时数据。

**Gemini**：很抱歉，我无法为您提供精确到2024年7月24日，四川成都市小麦种子价格、品种和规格的具体信息。

**ChatGPT**：在2024年7月24日，四川成都市的小麦种子价格和品种信息如下：品种：常见品种有"川麦"、"郑麦"、"南麦"等适合四川种植的优质小麦。规格：一般袋装，规格为50公斤一袋。价格：平均价格在200元至400元每袋不等，具体价格根据品种、质量和市场波动可能有所变化（China Seed Business）(National Bureau of Statistics)。如需更详细的供应商信息和实时报价，建议联系当地种子销售商或相关农资市场。

**WBLM**：科成麦4号，2.5公斤/袋，25元/袋。

Fig. 12. Examples of seed price query on different MLLMs.

## 5. Conclusion

In this study, we innovatively applied MLLM to wheat breeding and explored its potential for application in wheat breeding objectives. Based on different pre-trained open-source MLLMs, we used SFT, RAG, and RLHF technologies to inject cross-domain knowledge into MLLMs, thereby constructing multiple WBLMs. Among them, the WBLM constructed using SFT, RAG and RLHF technologies and InternVL2-8B has leading performance on the evaluation benchmark newly created in this study. The WBLM has competitive advantages over leading proprietary MLLMs, especially in wheat breeding tasks. Then, subsequent experiments were conducted using the WBLM. Ablation experiments indicated that the combination of SFT, RAG, and RLHF technologies can improve the overall generation performance, enhance the generated quality, balance the timeliness and adaptability of the generated answer, and reduce hallucinations and biases. The WBLM performed best in wheat yield prediction using cross-domain data (RS, phenotyping, weather, germplasm) simultaneously, with $R^2$ and RMSE of 0.821 and 489.254 kg/ha, respectively. Furthermore, the performance of WBLM and other MLLMs for wheat breeding goals was compared from a qualitative perspective of answer generation, and it was found that WBLM could generate more professional decision support answers. This study aims to provide an intelligent and integrated solution for wheat breeding goals, to help breeding work be carried out efficiently, to accelerate the breeding process of excellent varieties, and to provide scientific basis and technical support. In future, we will continue to explore deeper data fusion technology, optimize the generalization ability of the model, and enhance the interpretability of the model to ensure the transparency and reliability of breeding decisions. We will also expand the scope of application of the model to cover more crop varieties and environmental conditions to meet the breeding needs of different regions.

**CRediT authorship contribution statement**

**Guofeng Yang:** Methodology, Data curation, Formal analysis, Validation, Visualization, Writing – original draft. **Yu Li:** Writing – review & editing, Project administration. **Yong He:** Conceptualization, Project administration. **Zhenjiang Zhou:** Formal analysis, Writing – review & editing. **Lingzhen Ye:** Data curation, Project administration. **Hui Fang:** Data curation, Writing – review & editing. **Yiqi Luo:** Writing – review & editing. **Xuping Feng:** Conceptualization, Supervision, Funding acquisition, Writing - review & editing.

**Declaration of Competing Interest**

The authors declare that they have no known competing financial interests or personal relationships that could have appeared to influence the work reported in this paper.

**Acknowledgments**

Thanks to the Changxing Agricultural Experiment Station of Zhejiang University and the Yuhang Agricultural Experiment Station of Zhejiang University for providing the experimental base. Thanks to Professor Xianchun Xia and Dr. Jindong Liu (Chinese Academy of Agricultural Sciences) for providing wheat germplasm resources. Thanks to the members of the Digital Phenotyping Research Group of Zhejiang University for their help in data collection. This research was supported by Zhejiang Provincial Key R&D Program of China (Grant No. 2022C02013).

**Appendix A. Supplementary data**

Supplementary material

**Data Availability**

The dataset for the current study is available in the Zenodo repository at https://doi.org/10.5281/zenodo.13736773.